\documentclass[final]{cvpr}

\usepackage{times}
\usepackage{epsfig}
\usepackage{graphicx}
\usepackage{amsmath}
\usepackage{amssymb}
\usepackage{fixltx2e}
\usepackage{amsfonts}
\usepackage{listings}
\usepackage{algorithmic}
\usepackage{booktabs} 
\usepackage{multirow}
\usepackage{ amssymb } 
\usepackage{gensymb}
\usepackage{float}
\usepackage[labelfont=bf]{caption}
\usepackage[pagebackref=true,breaklinks=true,letterpaper=true,colorlinks,bookmarks=false]{hyperref}

\def\cvprPaperID{18} 
\def\confYear{CVPR 2021}

\begin{document}

\title{ASMNet: a Lightweight Deep Neural Network for Face Alignment and Pose Estimation }
\author{
Ali Pourramezan Fard, Hojjat Abdollahi, and Mohammad Mahoor\\
Department of Electrical and Computer Engineering\\
University of Denver, Denver, CO\\
{\tt\small $\left\{\text{Ali.pourramezanfard, hojjat.abdollahi, mohammad.mahoor}\right\}$@du.edu}
}

\maketitle


\begin{abstract}
Active Shape Model (ASM) is a statistical model of object shapes that represents a target structure. ASM can guide machine learning algorithms to fit a set of points representing an object (e.g., face) onto an image. This paper presents a lightweight Convolutional Neural Network (CNN) architecture with a loss function being assisted by ASM for face alignment and estimating head pose in the wild. We use ASM to first guide the network towards learning a smoother distribution of the facial landmark points. Inspired by transfer learning, during the training process, we gradually harden the regression problem and guide the network towards learning the original landmark points distribution. We define multi-tasks in our loss function that are responsible for detecting facial landmark points as well as estimating the face pose. Learning multiple correlated tasks simultaneously builds synergy and improves the performance of individual tasks. We compare the performance of our proposed model called ASMNet with MobileNetV2 (which is about 2 times bigger than ASMNet) in both the face alignment and pose estimation tasks. Experimental results on challenging datasets show that by using the proposed ASM assisted loss function, the ASMNet performance is comparable with MobileNetV2 in the face alignment task. In addition, for face pose estimation, ASMNet performs much better than MobileNetV2. ASMNet achieves an acceptable performance for facial landmark points detection and pose estimation while having a significantly smaller number of parameters and floating-point operations compared to many CNN-based models.
\end{abstract}


\section{Introduction}\label{sec:intro}

\begin{figure}[t]
  \centering
  \includegraphics[width=\columnwidth]{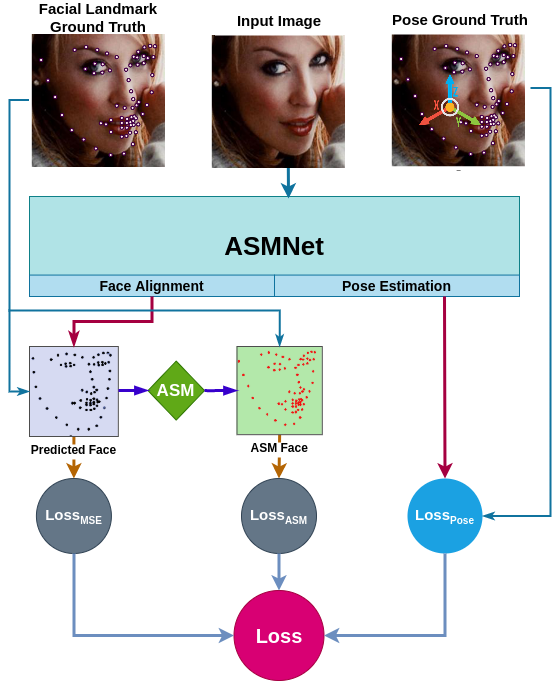}
  \caption{The proposed loss function ($Loss_{total}$) learns two main tasks simultaneously and uses ASM as assistant loss.}
  \label{fig:loss_function}
\end{figure}

Facial Landmark Points Detection is an essential task in many facial image analyses and applications. It is crucial for facial image alignment, face recognition~\cite{huang2013coupling, lu2015surpassing, soltanpour2017survey}, pose estimation~\cite{vicente2015driver}, and facial expression recognition~\cite{sun2014deep, zhao2003face}. Active Shape Model~(ASM) introduced by Tim Cootes~\cite{cootes1995active} is among the first methods designed for facial landmark points detection. ASM is a statistical shape model made out of object samples. ASM and its variant, Active Appearance Models (AAM)~\cite{cootes1998active, martins2013generative}, can guide learning algorithms to fit a set of points (e.g., facial points) representing an object into an image containing the object instance. In better words, ASM guides the learning algorithm to iteratively deforms the model to find the best match position between the model and the data in a new image. ASM/AAM and their predecessors' deformable models \cite{lucey2011} have been studied well for landmark point detection in facial image analysis and human body joint tracking. We propose to use ASM in a deep convolutional neural network (CNN) for facial landmark point detection and head pose estimation. 

Although most of the existing computer vision methods have focused on facial landmark points detection and pose estimation as separate tasks, some recent works \cite{chen2014joint,ranjan2017hyperface, zhang2015hierarchical,zhao2003face} show that learning correlated tasks simultaneously can improve the system accuracy. For instance, the authors of~\cite{zeiler2014visualizing} explain that information contained in the features is distributed throughout deep neural networks hierarchically. More specifically, while lower layers contain information about edges, and corners and hence are more appropriate for localization tasks such as facial landmark points detection and pose estimation, deeper layers contain more abstract information which is more suitable for classification tasks~\cite{ranjan2017hyperface}. Inspired by the idea of multi-task learning, we design our CNN model and hence the associated loss function to learn multiple correlated tasks simultaneously.

Several methods have been proposed for facial landmark points detection such as Constrained Local Model-Based Methods~\cite{asthana2013robust, cristinacce2006feature}, AAM~\cite{cootes1998active, martins2013generative}, part models \cite{zhuramanan2012}, and Deep Learning (DL) based methods~\cite{zhang2014coarse, zhang2014facial}. Although DL-based methods are considered as state-of-the-art methods, facial landmark points detection is still considered a challenging task specifically for faces with large pose variations~\cite{dong2018style, kumar2018disentangling, wu2018look}. Accordingly, the price to pay to achieve a high accuracy is the rise in computational complexity and the fall in efficiency. Recent methods have focused on improving the accuracy and this is normally achieved by introducing new layers and consequently increasing the number of parameters as well as longer inference time. These methods prove to be accurate and successful in desktop and server applications, but with the growth of IoT, mobile devices, and robotics, there is a growing need for more accurate and efficient algorithms. There are a few networks that are designed for low-power devices. One of the most popular ones is MobileNetV2~\cite{sandler2018mobilenetv2} which has proven to be a good feature extractor~\cite{huang2017speed}.

In this paper, we propose a new network structure that is inspired by MobileNetV2~\cite{sandler2018mobilenetv2} and is specifically designed for facial landmark points detection with the focus on making the network shallow and small without losing much accuracy. To achieve this goal we propose a new loss function that employs ASM as an assistant loss and uses multi-task learning to improve the accuracy. Fig.~\ref{fig:loss_function} depicts a general framework of our proposed idea. We tested our proposed method with the challenging 300W~\cite{sagonas2013300} dataset, and the Wider Facial Landmarks in the Wild (WFLW)~\cite{wu2018look} dataset. Our experimental results show that the accuracy of facial landmark points detection and pose estimation is comparable with the state-of-the-art methods while the size of the network is 2 times smaller than MobileNetV2~\cite{sandler2018mobilenetv2}. 

The remainder of this paper is organized as follows. Sec.~\ref{sec:lit} reviews the related work in facial landmark points detection, pose detection, and small neural networks. Sec.~\ref{sec:ASMNet} describes the architecture of our proposed network, the ASM assisted loss function and the training method. Experimental results are provided in Sec.~\ref{sec:experiment}. Finally, Sec.~\ref{sec:conclusion} concludes the paper with some discussions on the proposed method and future research directions.  

\section{Related Work}
\label{sec:lit}
Automated facial landmark points detection has been studied extensively by the computer vision community. Zhou \textit{etal.}~\cite{zhou2009automatic} classified facial landmark points detection methods into two categories: regression-based and template fitting methods. Regression-based methods consider a facial image as a vector and use a transformation such as Principle Component Analysis (PCA), Discrete Cosine Transform (DCT)~\cite{salah2007robust}, or Gabor Wavelets~\cite{arca2006face, vukadinovic2005fully} to transform the image into another domain. Then a classification algorithm such as SVM~\cite{antonini2003independent, du2008svm} or boosted cascade detector~\cite{viola2001robust} is used to detect facial landmarks. 
In contrast, template fitting methods such as Active Shape Models (ASM)~\cite{cootes1995active, ordas2003active} and Active Appearance Models (AAM)~\cite{cootes2004statistical} constrain the search for landmark positions by using prior knowledge. Inspired by the ASM method, we define a loss term that applies a constraint to the shapes learned during the training process.

Recently, Deep Learning techniques have dominated state-of-the-art results in terms of performance and robustness. There have been several new CNN-based methods for facial landmark points detection. Sun Y. \textit{etal.}~\cite{sun2013deep} proposed a deep CNN cascade to extract the facial key-points back in 2013. Zhang Z. \textit{etal.}~\cite{zhang2014facial} proposed a multi-task approach in which instead of solving FLP detection as an isolated task, they bundle it with similar tasks such as head pose estimation into a deep neural network to increase robustness and accuracy. Ranjan R. \textit{etal.}~\cite{ranjan2017hyperface} also uses deep multi-task learning and achieves high accuracy when detecting facial landmark. Several studies fit a 3D model onto the face and then infer the landmark positions~\cite{jourabloo2015pose, jourabloo2017pose, zhu2016face}.

One related task that can be trained simultaneously with facial landmark points detection, is head pose estimation. Detecting facial landmarks and estimating head pose can be made easier using 3D information~\cite{tulyakov2015regressing, zhu2016face}, however, this information is not always available. Wu~\textit{etal.}~\cite{wu2017simultaneous} propose a unified model for simultaneous facial landmark points detection, head pose estimation, and facial deformation analysis. This approach is robust to occlusion which is the result of the interaction between these tasks. One approach to estimate the head pose is to use the facial landmark point and head pose estimator sequentially~\cite{vicente2015driver}. We proposed a multi-task learning approach to tackle the problem of facial landmark points detection and pose estimation using a loss function assisted by ASM.

\section{Proposed ASM Network}
\label{sec:ASMNet}
We first review the Active Shape Model (ASM) algorithm and then introduce our proposed network architecture for landmark point localization and pose estimation. Finally, we explain our customized loss function based on ASM that improves the accuracy of the network. 

\subsection{Active Shape Model Review} \label{sec:asm_review}
Active Shape Model is a statistical model of shape objects. Each shape is represented as $n$ points, $S_{set}~=~\{ (S_x^1,S_y^1), ... ,(S_x^n,S_y^n)\}$ that are aligned into a common coordinate system. To simplify the problem and learn shape components, Principal Component Analysis (PCA) is applied to the covariance matrix calculated from a set of $K$ training shape samples. Once the model is built, an approximation of any training sample ($S$) is calculated using Eq.~\ref{eq: ASM_1}:
\begin{equation} \label{eq: ASM_1}
S \approx \overline{S}  + Pb 
\end{equation}
where $ \overline{S}$ is the sample mean,  $P = (p_{1}, p_{2}, ... , p_{t})$ contains $t$ eigenvectors of the covariance matrix and $b$ is a $t$ dimensional vector given by Eq.~\ref{eq: ASM_2}: 
\begin{equation} \label{eq: ASM_2}
b = P^\intercal (S - \overline{S})
\end{equation}
Consequently, a set of parameters of a deformable model is defined by vector $b$, so that by varying the elements of the vector, the shape of the model is changed. Consider that the statistical variance (\textit{i.e.}, eigenvalue) of the $i^{th}$ parameter of $b$ is $\lambda_i$. To make sure the generated image after applying ASM is relatively similar to the ground truth, the parameter $b_i$ of vector $b$ is usually limited to $\pm3\sqrt{\lambda_i}$~\cite{cootes2000introduction}. This constraint ensures that the generated shape is similar to those in the original training set. Hence, we create a new shape $S_{New}$ after applying this constraint, according to Eq.~\ref{eq:ASM_3}:
\begin{equation} \label{eq:ASM_3}
S_{New} = \overline{S}  + P\tilde{b} 
\end{equation}
where $\tilde{b}$ is the constrained $b$. We also define $\mathcal{ASM}$ operator according to Eq.~\ref{eq:ASM_4}:
\begin{equation} \label{eq:ASM_4}
\mathcal{ASM}:(P_x^i, P_y^i) \mapsto (A_x^i, A_y^i)
\end{equation}
$\mathcal{ASM}$ transforms each input point $(P_x^i, P_y^i)$ to a new point $(A_x^i, A_y^i)$ using Eqs.~\ref{eq: ASM_1}, \ref{eq: ASM_2}, and \ref{eq:ASM_3}.

\begin{figure}[t!]
  \centering
  \includegraphics[width=\columnwidth]{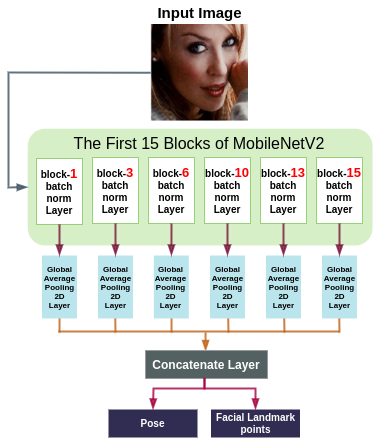}
  \caption{The architecture of the ASMNet network. ASMNet designed using first 15 blocks of MobileNetV2~\cite{sandler2018mobilenetv2}.}
  \label{fig:arch}
\end{figure}

In this paper, we propose a deep convolutional neural network architecture that utilizes ASM in the training loss function. Our proposed network (ASMNet) is significantly smaller than its predecessor, MobileNetV2~\cite{sandler2018mobilenetv2}, while its performance is comparable with MobileNetV2~\cite{sandler2018mobilenetv2} in localizing landmark points and much better in pose estimation.

\subsection{Proposed ASMNet Architecture} \label{sec:deepasmArch}
MobileNet and its variants \cite{sandler2018mobilenetv2} have received great attention as one of the most well-known deep neural networks for operation on embedded and mobile devices. Especially, MobileNetV2~\cite{sandler2018mobilenetv2} is shown to cope well with complex tasks such as image classification, object detection, and semantic segmentation. We have designed a network that is about two times smaller than MobileNetV2~\cite{sandler2018mobilenetv2}, both in terms of the number of parameters, and FLOPs. In designing ASMNet, we only use the first 15 blocks of MobileNetV2~\cite{sandler2018mobilenetv2} while the main architecture which has 16 blocks. Nevertheless, creating a shallow network would eventually lower the final accuracy of the system. To avoid this problem we purposefully add a few new layers. Fig.~\ref{fig:arch} shows the architecture of ASMNet.

According to~\cite{zeiler2014visualizing} the features in a Convolutional Neural Network (CNN) are distributed hierarchically. In other words, lower layers have features such as edges, and corners which are more suitable for tasks like landmark localization and pose estimation, and deeper layers contain more abstract features that are more suitable for tasks like image classification and image detection. Training a network for correlated tasks simultaneously builds a synergy that can improve the performance of each task~\cite{chen2014joint, zhang2014facial}.

Motivated by the approach in \cite{ranjan2017hyperface}, we designed a multi-task CNN to detect facial landmarks as well as estimating the pose of the faces (pitch, roll, and yaw) simultaneously. 
In order to use features from different layers, we have created shortcuts from \textit{block-1-batch-normalization}, \textit{block-3-batch-normalization}, \textit{block-6-batch-normalization}, \textit{block-10-batch-normalization}, and finally \textit{block-13-batch-normalization}. We connect each of these shortcuts to the output of block 15 of MobileNetV2~\cite{sandler2018mobilenetv2}, \textit{block-15-add}, using a \textit{global average pooling} layer. Finally, we concatenate all the \textit{global average pooling} layers. Such architecture enables us to use features that are available in different layers of the network while keeping the number of the FLOPs small. In other words, since the original MobileNetV2~\cite{sandler2018mobilenetv2} is designed for image classification task - where the more abstract features are required - it might not be suitable for face alignment task - which needs both abstract features that are available in the deeper layers as well as features that are available in the lower layers such as edges and corners -. 

We Designed ASMNet (see Fig.~\ref{fig:arch}), by fusing the features that are available if different layers of the model. Furthermore, by concatenating the features that are collected after each \textit{global average pooling} layer in the back-propagation process, it will be possible for the network to evaluate the effect of each shortcut path.

Moreover, we add another correlated task to the network. As Fig.~\ref{fig:arch} shows, the proposed network predicts 2 different outputs: the facial landmark points (the main output of the network), as well as the face pose. While the correlation and the synergy between these two tasks can result in more accurate results, we also wanted our lightweight ASMNet to be able to predict face pose as well so that it might be used in more applications.

\subsection{ASM Assisted Loss Function} \label{sec:asmloss}
In the following, we describe the loss functions for two different tasks. These tasks are responsible for \textit{facial landmark points detection}, and \textit{pose estimation}.

\textbf{Facial landmark points detection task:}~The common loss function for facial landmark points detection is Mean Square Error (MSE). We propose a new loss function that including MSE, as the main loss as well an the \textit{assistant} loss which utilizes ASM to improve the accuracy of the network called \textit{ASM-LOSS}. 

The proposed ASM-LOSS guides the network to first learn the smoothed distribution of the facial landmark points. In other words, during the training process, the loss function compares the predicted facial landmark points with their corresponding ground truth as well as the smoothed version the ground truth which is generated using ASM. Given this, in the early stage of training, we set a bigger weight to the ASM-LOSS in comparison to the main loss -- which is MSE --, since the variation of the smoothed facial landmark points are much lower than the original landmark points, and as a rule of thumb, easier to be learned by a CNN. Then, by gradually decrease the weight of the ASM-LOSS, we lead the network to focus more on the original landmark points. In practice, we figured out that this method, which is also can be taken to account as transfer learning, works out well and results in more accurate models.

We also discover that although face pose estimation has a heavy reliance on face alignment, it can achieve good accuracy with the assistant of smoothed facial landmark points as well. In other words, if the performance of facial landmark point detection task is  \textit{acceptable}, which means network can predict facial landmark such that the whole shape of the face is correct, the pose estimation can achieve a good accuracy. Accordingly, using smoothed landmark points and training network using ASM-LOSS will results in a more accuracy in pose estimation task.

Consider that for each image in the training set, there exists \textit{n} landmark points in a set called \textit{G} such that $(G_x^i,G_y^i)$ is the coordinates for the $i^{th}$ landmark point. Similarly, a predicted set \textit{P} contains \textit{n} points such that $(P_x^i,P_y^i)$ is the predicted coordinates for the $i^{th}$ landmark point.

\begin{equation}
\begin{split}
G_{set}  = \{ (G_x^1,G_y^1), ... ,(G_x^n,G_y^n)\}\\
P_{set}  = \{ (P_x^1,P_y^1), ... ,(P_x^n,P_y^n)\}
\end{split}
\end{equation}

We apply PCA on the training set and calculate \textit{eigenvectors} and \textit{eigenvalues}. Then, we calculate set \textit{A}, which contains \textit{n} points and each point is the transformation of the corresponding point in \textit{G}, by applying the $\mathcal{ASM}$ operator according to Eq.~\ref{eq:ASM_4}:
\begin{equation}
\begin{split}
A_{set}  = \{(A_x^1,A_y^1), ... ,(A_x^n,A_y^n)\}\\
\mathcal{ASM}:(G_x^i, G_y^i) \mapsto (A_x^i, A_y^i)
\end{split}
\end{equation}

We define the \textit{main} facial landmark point loss, Eq.~\ref{eq:mse_loss}, as the Mean Square Error between the ground truth (\textit{G}) and the predicted landmark points (\textit{P}). 
\begin{equation} \label{eq:mse_loss}
    \mathcal{L}_{mse} = \frac{1}{N} \frac{1}{n} \sum_{j=1}^{N} \sum_{i=1}^{n} \lVert G^i_j - P^i_j \rVert_2 
\end{equation}
where $N$ is the total number of images in the training set and $G^i_j = (G_x^i, G_y^i)$ shows the $i^{th}$ landmark of the $j^{th}$ sample in the training set. We calculate ASM-LOSS as the error between ASM points ($A_{set}$), and predicted landmark points ($P_{set}$) using Eq.~\ref{eq:asm_loss}:
\begin{equation} \label{eq:asm_loss}
    \mathcal{L}_{asm} = \frac{1}{N} \frac{1}{n} \sum_{j=1}^{N}  \sum_{i=1}^{n} \lVert A^i_j - P^i_j \rVert_2
\end{equation}

Finally, we calculate the \textit{total} loss for the \textit{facial landmark task} with according to Eq.~\ref{eq:facial_total_loss}:
\begin{equation} \label{eq:facial_total_loss}
     \mathcal{L}_{facial} = \mathcal{L}_{mse} + \alpha~\times~\mathcal{L}_{asm}
\end{equation}
The accuracy of PCA have a heavy reliance on the ASM points ($A_{set}$), which means that the more accurate the PCA, the less the discrepancy between the ground truth (\textit{G}) and the ASM points ($A_{set}$). To be more detailed, by reducing the accuracy of PCA, the generated ASM points ($A_{set}$), will be more similar to the \textit{average point set}, which is the average of all the ground truth face objects in the training sets. Consequently, predicting points in $A_{set}$ is easier than the points in the $G_{set}$ since the variation of latter is lower than the variation of the former. We use this feature to design our loss function such that we first guide the network towards learning the distribution of the smoothed landmark points -- which is easier to be learned -- and gradually harden the problem by decreasing the weight of ASM-LOSS.

We define $\alpha$ as ASM-LOSS weight using Eq.~\ref{eq:alpha_weight}:
\begin{equation} \label{eq:alpha_weight}
    \alpha = \left\{
        \begin{matrix} 2 && i < \frac{l}{3}
        \\ 1    && \frac{l}{3} < i < \frac{2l}{3}
        \\ 0.5  && i > \frac{2l}{3}
        
        \end{matrix}\right.
\end{equation}
where  \textit{i} is the epoch number and \textit{l} is the total number of training epochs. As shown in Eqs.~\ref{eq:facial_total_loss}, at the beginning of the training, the value of $\alpha$ is higher, which means we put more emphasize on ASM-LOSS. Hence, the network focuses more on predicting a simpler task and converges faster. Then after one-third of total epochs, we reduce $\alpha$ to 1, and put equal emphasis on the main MSE loss ASM-LOSS. Finally, after two-third of total epochs, by reducing $\alpha$ to 0.5, we direct the network toward predicting the main ground truths, while considering the smoothed points generated using ASM as an assistant. We also show experimentally in Sec.~\ref{sec:experiment}, that such technique leads to more accurate results, specifically when it comes to a lightweight network like ASMNet.

\textbf{Pose estimation task:}~We use mean square error to calculate the loss for the head pose estimation task. Eq.~\ref{eq:pose_loss} defines the loss function \textit{$\mathcal{L}_{pose}$}, where yaw(\textit{$y^{p}$}), pitch(\textit{$p^{p}$}), and roll(\textit{$r^{p}$}) are the predicted poses and \textit{$y^{t}$}, \textit{$p^{t}$}, and \textit{$r^{t}$} are the corresponding ground truths.
\begin{equation} \label{eq:pose_loss}
     \mathcal{L}_{pose} = \frac{1}{N} \sum_{j=1}^{N} \frac{ (y^{p}_j - y^{t}_j)^2 + (p^{p}_j - p^{t}_j)^2 + (r^{p}_j - r^{t}_j)^2 }{3} 
\end{equation}

Finally, we calculate the total loss as the total weighted loss of the 2 individual losses using Eq.~\ref{eq:total_loss}:
\begin{equation} \label{eq:total_loss}
     \mathcal{L} = \sum_{i=1}^{2} \lambda_{task_i} \mathcal{L}_{task_i}
\end{equation}
such that \textit{$task_i$} is the $i^{th}$ element of the task set T~=~\{~$\mathcal{L}_{facial}$, $\mathcal{L}_{pose}$~\} and the value of $\lambda_{task_i}$ corresponds to the importance of the $i^{th}$ task. Since we define facial landmark points detection task to be more important than pose estimation, we choose $\lambda_{task} =\{1, 0.5\}$. Fig.~\ref{fig:loss_function} illustrates the process of calculating the total loss value.

\section{Experimental Results}
\label{sec:experiment}

\subsection{Training Phase} \label{sec:training}


\textbf{300W}.We followed the protocol described in~\cite{ren2014face} to train our networks on the 300W~\cite{sagonas2013300} dataset. We use 3,148 faces consisting of 2,000 images from the training subset of HELEN~\cite{le2012interactive} dataset, 811 images from the training subset of LFPW~\cite{belhumeur2013localizing} dataset, and 337 images from the full set of AFW~\cite{zhuramanan2012} dataset with a 68-point annotation. For testing, 300W~\cite{sagonas2013300} has 3 subsets: Common subset with 554 images, Challenging subset with 135 images, and Full subset, including both Common and Challenging subsets, with 689 images. More specifically, the Challenging subset is the IBUG~\cite{sagonas2013300} dataset while the Common subset is a combination of the HELEN test subset (330 images) and LFPW test subset (224 images).


\textbf{WFLW}. WFLW~\cite{wu2018look}, containing 7500 images for training and 2500 images for testing, is another widely used dataset, recently has been proposed based on WIDER FACE~\cite{yang2016wider}. Each image in this dataset contains 98 manual annotated landmarks. In order to be able to evaluate the models under different circumstances, WFLW~\cite{wu2018look} provides 6 different subsets including 314 expression images, 326 large pose images, 206 make-up images, 736 occlusion images, 698 illumination images, and 773 blur images. 



We use the method and algorithm in~\cite{Ruiz_2018_CVPR_Workshops} to calculate the yaw, roll, and, pitch for each image in the dataset since to the best of our knowledge, no dataset provides the annotation for facial landmark points and face pose jointly.


\subsection{Implementation Details}
For the training set in each dataset, we crop all the images and extract the face region. Then the face images are scaled to $224\times224$ pixels. We augment the images (in terms of contrast, brightness, and color) to add robustness of data variation to the network. We use Adam optimizer for training the networks with learning rate $10^{-2}$, $\beta_1 = 0.9$, $\beta_2 = 0.999$, and $decay = 10^{-5}$. Then we train networks for about 150 epochs with a batch size of 50. We implemented our codes using the TensorFlow library and run them on a NVidia 1080Ti GPU.

\subsection{Evaluation Metrics} \label{sec:vel_metrics}
We follow the previous works and employ normalized mean error~(NME) to measure the accuracy of our model. We define the normalising factor, followed by MDM~\cite{trigeorgis2016mnemonic} and~\cite{sagonas2013300} as “inter-ocular” distance (the distance between the outer-eye-corners). Furthermore, we calculate failure rate~(FR), defined as the proportion of failed detected faces, for a maximum error of 0.1. Cumulative Errors Distribution~(CED) curve as well as the area-under-the-curve~(AUC)~\cite{yang2015empirical} is also reported. Besides, we use mean absolute error~(MAE) for evaluating the pose estimation task.



\subsection{Comparison with Other Models} \label{sec:result_comparison}

We conducted four different experiments to evaluate the effectiveness of the proposed ASM assisted loss function. These experiments are designed to assess the performance of MobileNetV2~\cite{sandler2018mobilenetv2} and ASMNet, with and without the proposed ASM assisted loss function.

Table~\ref{tbl:tbl_internal_comparison} shows the results of the experiments on \textit{Full} subsets of 300W~\cite{sagonas2013300} (full), and WFLW~\cite{wu2018look}, as well as the number of network parameters(\#Params) and the sum of the FLOPs. For simplicity, we name our model as "mnv2" (MobileNetV2~\cite{sandler2018mobilenetv2} trained using standard MSE loss function), "mnv2\_{r}" (MobileNetV2~\cite{sandler2018mobilenetv2} trained using our ASM assisted loss function), "ASMNet\_{nr}" (ASMNet trained using standard MSE loss function), and "ASMNet" ( ASMNet trained using our ASM assisted loss function).

Table~\ref{tbl:tbl_internal_comparison} shows that the proposed ASM assisted loss function has a lower NME in both cases. Furthermore, while our proposed network architecture is about two times smaller than MobileNetV2~\cite{sandler2018mobilenetv2}, its performance is comparable with it after applying our proposed ASM assisted loss function. It means that without sacrificing accuracy, we have created a network that is smaller and faster in comparison to MobileNetV2~\cite{sandler2018mobilenetv2}. Such characteristics make the ASMNet suitable for running on mobile and embedded devices.

\begin{table}[b]
\centering
\caption{Number of parameters in million (M) and FLOPs in billion (B), as well as Normalized Mean Error (NME in \%) of landmarks localization on 300W~\cite{sagonas2013300}, and WFLW~\cite{wu2018look} datasets.}
\label{tbl:tbl_internal_comparison}
\resizebox{6cm}{!}{

\begin{tabular}{@{}lcccc@{}}
\toprule
\multirow{2}{*}{Method} & \multicolumn{2}{c}{NME} & \multirow{2}{*}{Params (M)} & \multirow{2}{*}{FLOPs (B)} \\ \cmidrule(lr){2-3}
                        & 300W       & WFLW       &                             &                            \\ \midrule
mnv2                    & 4.70       & 9.57       & \multirow{2}{*}{2.42}       & \multirow{2}{*}{0.60}      \\
mnv2\_r             & 4.59       & 9.41       &                             &                            \\
ASMNet\_nr          & 6.49       & 11.96      & \multirow{2}{*}{1.43}       & \multirow{2}{*}{0.51}      \\
ASMNet                  & 5.50       & 10.77      &                             &                            \\ \bottomrule
\end{tabular}

}
\end{table}

\begin{figure}[t]
  \centering
  \includegraphics[width=\columnwidth]{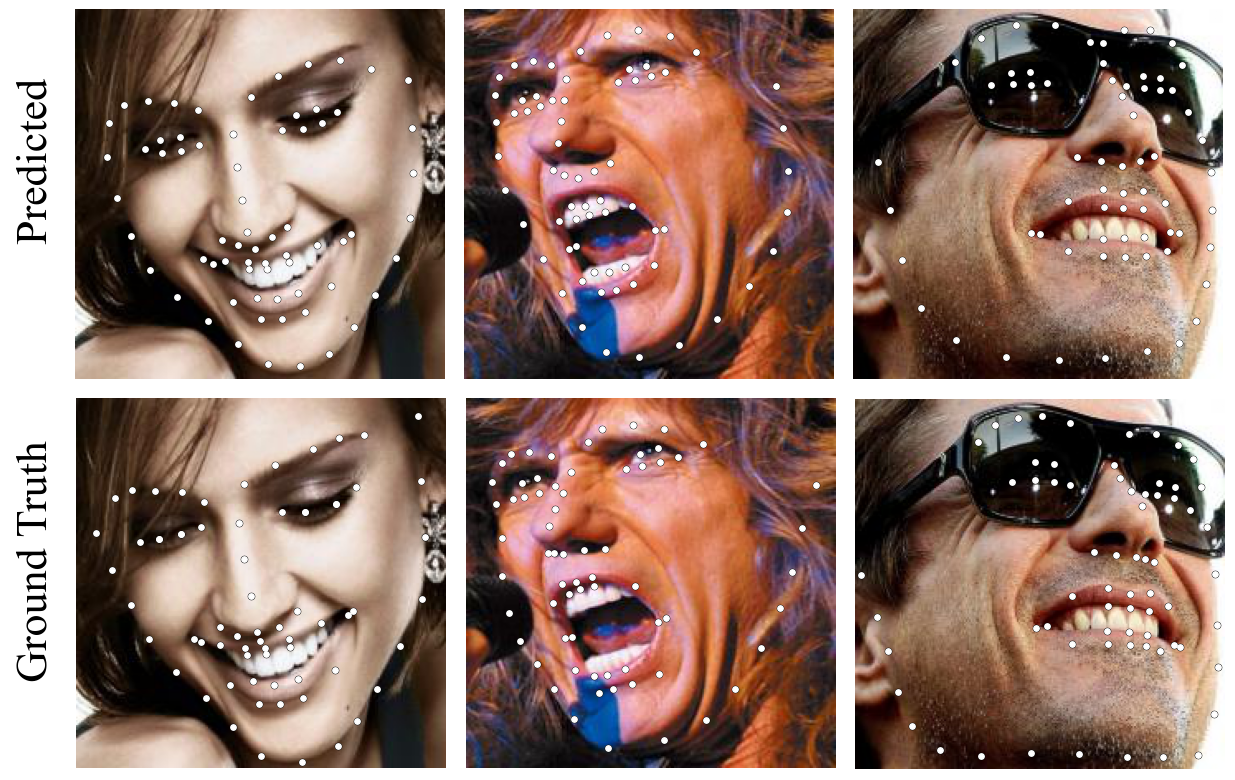}
  \caption{Facial Landmark detection using ASMNet over 300W~\cite{sagonas2013300} Challenging subset.}
  \label{fig:result_300w}
\end{figure}

\textbf{Evaluation on 300W.}
The 300W~\cite{sagonas2013300}  dataset is a very challenging benchmark in facial landmark detection task. Table~\ref{tbl:tbl_results_300w} shows a comparison between ASMNet and the state-of-the-art methods. Although the performance of ASMNet does not outperform the state-of-the-art methods, comparing the number of the parameters, and FLOPs of the models (see Table~\ref{tbl:size_cost_analysis}), the accuracy of our proposed model is comparable and accurate in the context of small networks such as MobileNetV2~\cite{sandler2018mobilenetv2}. Furthermore, As the table~\ref{tbl:tbl_results_300w} shows, the performance of the ASMNet with ASM assisted loss function on 300W~\cite{sagonas2013300} is better than the performance of ASMNet without the assisted loss. Fig.~\ref{fig:result_300w} shows the some example of facial landmark detection using ASMNet on the Challenging subset of 300W~\cite{sagonas2013300} dataset. As we can see, ASMNet performs well, even in challenging face images.

\begin{table}[t] 
\caption{Normalized Mean Error (in \%) of 68-point landmarks localization on 300W~\cite{sagonas2013300} dataset.}
\label{tbl:tbl_results_300w}
\centering
\small
\resizebox{6cm}{!}
{\begin{tabular}{l c c c  }
\hline
\multirow{2}{*}{Method} & \multicolumn{3}{c }{Normalized Mean Error}       \\ 
                                 & Common & Challenging  & Fullset \\ \hline
RCN~\cite{honari2016recombinator}    & 4.67           & 8.44        & 5.41             \\ 
DAN~\cite{kowalski2017deep}    & 3.19           & 5.24        & 3.59             \\
PCD-CNN~\cite{kumar2018disentangling}    & 3.67           & 7.62        & 4.44             \\ 
CPM~\cite{dong2018supervision}    & 3.39           & 8.14        & 4.36             \\ 
DSRN~\cite{miao2018direct}    & 4.12           & 9.68        & 5.21             \\ 
SAN~\cite{dong2018style}    & 3.34           & 6.60        & 3.98             \\ 
LAB~\cite{wu2018look}    & 2.98           & 5.19        & 3.49             \\ 
DCFE~\cite{valle2018deeply}           & 2.76            & 5.22        & 3.24             \\ \hline 

mnv2                 &  3.93            & 7.52       & 4.70            \\ 
mnv2\_{r}           & 3.88            & 7.35        & 4.59           \\ 
ASMNet\_{nr}     &5.86     & 8.80 & 6.46       \\ 
ASMNet          &4.82     & 8.2 & 5.50       \\ \hline 

\end{tabular}}
\end{table}

\begin{table*}[]
\caption{Normalized Mean Error (in \%), failure rate (in \%), and AUC of 98-point landmarks localization on WFLW~\cite{wu2018look} dataset.}
\label{tbl:tbl_results_wflw}
\centering
\small
\resizebox{16cm}{!}
{
\begin{tabular}{ l l p{1.4cm} p{1.4cm} p{1.4cm} p{1.5cm} p{1.4cm} p{1.4cm} p{1.4cm}}
\hline
Metric                  & Method                                                                            & Test set  & Pose  & Expression & Illumination    & Make-Up   & Occlusion    & Blur                                                                              \\ \hline 
\multirow{2}{*}{\rotatebox[origin=c]{90}{Mean Error (\%)}} & \begin{tabular}[c]{@{}l@{}}
ESR~\cite{cao2014face}\\ SDM~\cite{xiong2013supervised}\\ CFSS~\cite{zhu2015face}\\ DVLN~\cite{wu2017leveraging}\\ LAB~\cite{wu2018look}\\ ResNet50(Wing+PDB)~\cite{feng2018wing}
\end{tabular} & \begin{tabular}[c]{@{}l@{}}11.13\\ 10.29\\ 9.07\\ 6.08\\ 5.27\\ 5.11\end{tabular}  & \begin{tabular}[c]{@{}l@{}}25.88\\ 24.10\\ 21.36\\ 11.54\\ 10.24\\ 8.75\end{tabular}             & \begin{tabular}[c]{@{}l@{}}11.47\\ 11.45\\ 10.09\\ 6.78\\ 5.51\\ 5.36\end{tabular}               & \begin{tabular}[c]{@{}l@{}}10.49\\ 9.32\\ 8.30\\ 5.73\\ 5.23\\ 4.93\end{tabular}                 & \begin{tabular}[c]{@{}l@{}}11.05 \\ 9.38 \\ 8.74 \\ 5.98 \\ 5.15 \\ 5.41\end{tabular}            & \begin{tabular}[c]{@{}l@{}}13.75\\ 13.03\\ 11.76\\ 7.33\\ 6.79\\ 6.37\end{tabular}               & \begin{tabular}[c]{@{}l@{}}12.20\\ 11.28\\ 9.96\\ 6.88\\ 6.32\\ 5.81\end{tabular}                \\ \cline{2-9} 
 & \begin{tabular}[c]{@{}l@{}}mnv2\\ mnv2\_r\\ ASMNet\_nr\\ ASMNet\end{tabular} 
 & \begin{tabular}[c]{@{}l@{}} 9.57\\ 9.41\\ 11.96\\ 10.77\end{tabular}
 & \begin{tabular}[c]{@{}l@{}} 18.18\\ 17.86\\ 21.95\\ 21.11\end{tabular}                                        
 & \begin{tabular}[c]{@{}l@{}} 9.93\\ 9.78\\ 13.08\\ 12.02\end{tabular}
 & \begin{tabular}[c]{@{}l@{}} 8.98\\  8.90\\ 11.02\\  9.93\end{tabular}                                            
 & \begin{tabular}[c]{@{}l@{}} 9.92\\ 9.67\\ 11.84\\ 10.55\end{tabular}                                            
 & \begin{tabular}[c]{@{}l@{}} 11.38\\ 11.25\\ 13.24\\ 12.34\end{tabular}                                            
 & \begin{tabular}[c]{@{}l@{}} 10.79\\ 10.66\\12.60\\ 11.62\end{tabular}                                            \\ \hline

\multirow{2}{*}{\rotatebox[origin=c]{90}{Failure Rate}}    & \begin{tabular}[c]{@{}l@{}}
ESR~\cite{cao2014face}\\ SDM~\cite{xiong2013supervised}\\ CFSS~\cite{zhu2015face}\\ DVLN~\cite{wu2017leveraging}\\ LAB~\cite{wu2018look}\\ ResNet50(Wing+PDB)~\cite{feng2018wing}
\end{tabular} & \begin{tabular}[c]{@{}l@{}}35.24\\ 29.40\\ 20.56\\ 10.84\\ 7.56\\ 6.00\end{tabular}            & \begin{tabular}[c]{@{}l@{}}90.18\\ 84.36\\ 66.26\\ 46.93\\ 28.83\\ 22.70\end{tabular}            & \begin{tabular}[c]{@{}l@{}}42.04\\ 33.44\\ 23.25\\ 11.15\\ 6.37 \\ 4.78\end{tabular}             & \begin{tabular}[c]{@{}l@{}}30.80 \\ 26.22 \\ 17.34 \\ 7.31 \\ 6.73 \\ 4.30\end{tabular}          & \begin{tabular}[c]{@{}l@{}}38.84 \\ 27.67 \\ 21.84 \\ 11.65 \\ 7.77 \\ 7.77\end{tabular}         & \begin{tabular}[c]{@{}l@{}}47.28 \\ 41.85 \\ 32.88 \\ 16.30 \\ 13.72 \\ 12.50\end{tabular}       & \begin{tabular}[c]{@{}l@{}}41.40 \\ 35.32 \\ 23.67 \\ 13.71 \\ 10.74 \\ 7.76\end{tabular}        \\ \cline{2-9} 
 & \begin{tabular}[c]{@{}l@{}}mnv2\\ mnv2\_r\\ ASMNet\_nr\\ ASMNet\end{tabular} 
 & \begin{tabular}[c]{@{}l@{}} 30.64\\ 30.04\\ 50.2\\ 39.12\end{tabular}                                          
 & \begin{tabular}[c]{@{}l@{}}  88.03\\ 88.65\\ 98.46\\ 98.41\end{tabular}                                    
 & \begin{tabular}[c]{@{}l@{}}  34.07\\  31.52\\ 70.38\\ 59.87\end{tabular}                                             
 & \begin{tabular}[c]{@{}l@{}} 25.39\\  24.67\\ 43.68\\ 33.38\end{tabular}                                            
 & \begin{tabular}[c]{@{}l@{}} 32.03\\ 30.09\\ 50.0\\ 38.34\end{tabular}                                            
 & \begin{tabular}[c]{@{}l@{}}  41.84\\ 41.44\\ 59.78\\ 48.64\end{tabular}                                    
 & \begin{tabular}[c]{@{}l@{}}  38.80\\ 37.25\\ 56.14\\  46.31\end{tabular}                                            \\ \hline
\multirow{2}{*}{\rotatebox[origin=c]{90}{AUC}}             & \begin{tabular}[c]{@{}l@{}}
ESR~\cite{cao2014face}\\ SDM~\cite{xiong2013supervised}\\ CFSS~\cite{zhu2015face}\\ DVLN~\cite{wu2017leveraging}\\ LAB~\cite{wu2018look}\\ ResNet50(Wing+PDB)~\cite{feng2018wing}
\end{tabular} & \begin{tabular}[c]{@{}l@{}}0.2774\\ 0.3002\\ 0.3659 \\ 0.4551 \\ 0.5323 \\ 0.5504\end{tabular} & \begin{tabular}[c]{@{}l@{}}0.0177 \\ 0.0226 \\ 0.0632 \\ 0.1474 \\ 0.2345 \\ 0.3100\end{tabular} & \begin{tabular}[c]{@{}l@{}}0.1981 \\ 0.2293 \\ 0.3157 \\ 0.3889 \\ 0.4951 \\ 0.4959\end{tabular} & \begin{tabular}[c]{@{}l@{}}0.2953 \\ 0.3237 \\ 0.3854 \\ 0.4743 \\ 0.5433 \\ 0.5408\end{tabular} & \begin{tabular}[c]{@{}l@{}}0.2485 \\ 0.3125 \\ 0.3691 \\ 0.4494 \\ 0.5394 \\ 0.5582\end{tabular} & \begin{tabular}[c]{@{}l@{}}0.1946 \\ 0.2060 \\ 0.2688 \\ 0.3794 \\ 0.4490 \\ 0.4885\end{tabular} & \begin{tabular}[c]{@{}l@{}}0.2204 \\ 0.2398 \\ 0.3037 \\ 0.3973 \\ 0.4630 \\ 0.4918\end{tabular} \\ \cline{2-9} 
& \begin{tabular}[c]{@{}l@{}}mnv2\\ mnv2\_reg\\ ASMNet\_nr\\ ASMNet\end{tabular} 
&\begin{tabular}[c]{@{}l@{}}    0.2388\\ 0.2447\\ 0.1024\\ 0.1637 \end{tabular} 
& \begin{tabular}[c]{@{}l@{}}   0.0096\\ 0.0099\\ 0.0008\\ 0.0010 \end{tabular} 
&\begin{tabular}[c]{@{}l@{}}    0.1812\\ 0.1836\\ 0.0414\\ 0.0714 \end{tabular} 
&\begin{tabular}[c]{@{}l@{}}    0.2510\\ 0.2563\\ 0.1129\\ 0.1826 \end{tabular}
& \begin{tabular}[c]{@{}l@{}}   0.2147\\ 0.2282\\ 0.0941\\ 0.1653 \end{tabular} 
&\begin{tabular}[c]{@{}l@{}}    0.1719\\ 0.1779\\ 0.0729\\ 0.1202 \end{tabular}
&\begin{tabular}[c]{@{}l@{}}    0.1852\\ 0.1880\\ 0.0797\\ 0.1268 \end{tabular}                                            
\\ \hline
\end{tabular}}
\end{table*}

\begin{figure}[t!]
  \centering
  \includegraphics[width=\columnwidth]{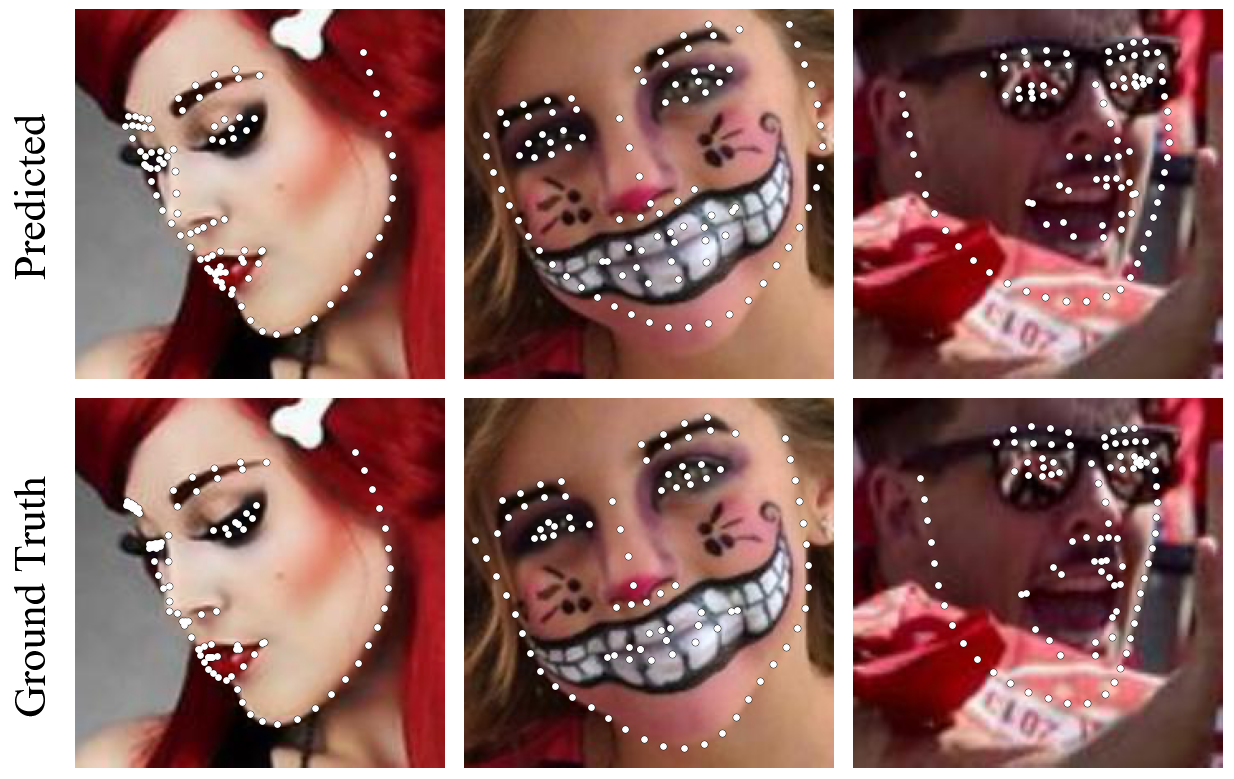}
  \caption{Facial Landmark detection using ASMNet over WFLW~\cite{wu2018look} dataset.}
  \label{fig:result_WFLW}
\end{figure}

\textbf{Evaluation on WFLW.}
Table~\ref{tbl:tbl_results_wflw} shows the performance of the state-of-the-art method and our proposed method over WFLW~\cite{wu2018look} and its 6 subsets. The performance of ASMNet is comparable to the performance of MobileNetV2~\cite{sandler2018mobilenetv2}. In other words, using the proposed ASM assisted loss function improves the model accuracy. Fig.~\ref{fig:result_WFLW} shows the some example of facial landmark detection using ASMNet on WFLW~\cite{wu2018look} dataset. While ASMNet can be taken as a very lightweight model, its performance is acceptable under different circumstances such as occlusion, extreme pose, expression, illumination, blur, and make-up.

\begin{table}[t]
\caption{Mean Absolute Error of pose estimation on 300W~\cite{sagonas2013300}, WFLW~\cite{wu2018look} datasets compared to HopeNet\cite{Ruiz_2018_CVPR_Workshops}.}
\label{tbl:pose_estimation_result}
\centering
\small
\resizebox{7cm}{!}
{\begin{tabular}{l c c c c c }
\hline
\multicolumn{2}{c}{Method} & \multicolumn{1}{c }{ASMNet\_nr} & \multicolumn{1}{c }{ASMNet} & \multicolumn{1}{c}{mnv2} & \multicolumn{1}{c}{mnv2\_r} \\ \hline 
\multirow{3}{*}{300W~\cite{sagonas2013300}}
                      & yaw   & 2.41  & 1.62 & 1.75 & 1.71 \\ 
                      & pitch & 1.87  & 1.80 & 1.93 & 1.89 \\ 
                      & roll  & 2.115 & 1.24 & 1.32 & 1.30 \\ \hline
\multirow{3}{*}{WFLW~\cite{wu2018look}} 
                      & yaw   & 3.14  & 2.97 & 3.06 & 3.08 \\ 
                      & pitch & 2.99  & 2.93 & 3.03 & 2.94 \\ 
                      & roll  & 2.23  & 2.21 & 2.26 & 2.22 \\ \hline
\end{tabular}}
\end{table}

\begin{table}[t]
\caption{Mean Absolute Error of pose estimation on using ASMNet, JFA~\cite{xu2017joint}, and Yang\textit{et. al}~\cite{yang2015face} on 300W~\cite{sagonas2013300}.}
\label{tbl:pose_estimation_result_external}
\centering
\small

\begin{tabular}{l c c c }
\hline
Method & Pitch & Yaw  & Roll \\
\hline

Yang\textit{et. al}~\cite{yang2015face}   & 5.1   & 4.2  & 2.4  \\
JFA~\cite{xu2017joint}                  & 3.0   & 2.5  & 2.6  \\ \hline
ASMNet                                  & 1.80  & 1.62 & 1.24 \\ \hline
\end{tabular}
\end{table}

\textbf{Pose Evaluation.}
Neither 300W~\cite{sagonas2013300} nor WFLW~\cite{wu2018look} dataset do not provide the head pose information. Accordingly, we followed the method used by~\cite{xu2017joint} and used another application to synthesizes the pose information. Although~\cite{xu2017joint} used \cite{asthana2013robust} for synthesizing the pose information, we used HopeNet~\cite{Ruiz_2018_CVPR_Workshops} which is a state-of-the-art pose estimation method. Using HopeNet we acquired the \textit{yaw}, \textit{pitch}, and \textit{roll} values of the 300W~\cite{sagonas2013300}, and WFLW~\cite{wu2018look} images and used them as the ground truths for our network. Table~\ref{tbl:pose_estimation_result} shows the mean absolute error~(MAE) between HopeNet~\cite{Ruiz_2018_CVPR_Workshops} results and our ASMNet. 

In addition, we compare the performance of our proposed method with~\cite{xu2017joint} as well as~\cite{yang2015face} in Table~\ref{tbl:pose_estimation_result_external} using \textit{Full} subset of 300W~\cite{sagonas2013300} dataset. As the results show, the performance of our lightweight ASMNet is comparable to HopeNet~\cite{Ruiz_2018_CVPR_Workshops}, which is a state-of-the-art method and outperforms the other methods as well. Besides, the performance of ASMNet is better than MobileNetV2~\cite{sandler2018mobilenetv2}, even when it utilizes the ASM-LOSS function. Since in pose estimation task aligning the whole shape of the face is more crucial than aligning each landmark point, using ASM-LOSS function will lead to better performance. 

\begin{figure}[t]
  \centering
  \includegraphics[width=\columnwidth]{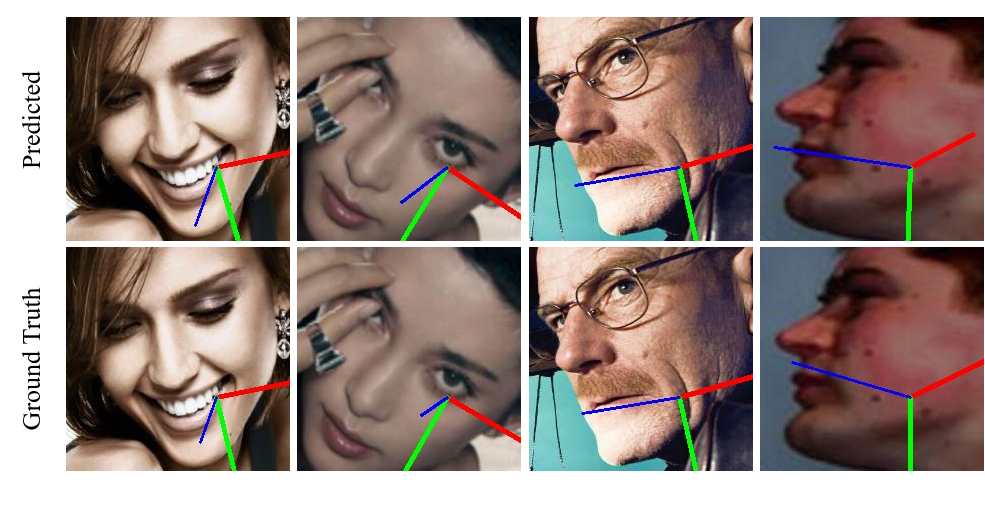}
  \caption{ASMNet can also estimate the head pose even in challenging conditions. The input images are from 300W~\cite{sagonas2013300} Challenging set.}
  \label{fig:output_pose}
\end{figure}

Moreover, ASMNet is designed to use features generated in different layers of the neural network which enables it to outperforms MobileNetV2~\cite{sandler2018mobilenetv2} in pose estimation task. Fig.~\ref{fig:output_pose} shows the output of the pose estimation task.

\begin{figure}[t!]
  \centering
  \includegraphics[width=\columnwidth]{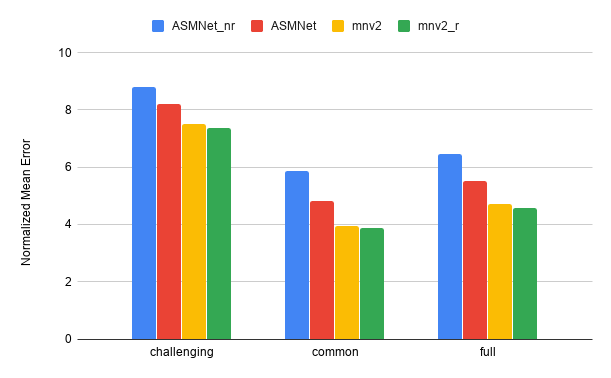}
  \caption{Comparing the performance of ASMNet, as well as MobileNetV2~\cite{sandler2018mobilenetv2} with and without the proposed ASM assisted loss function on 300W~\cite{sagonas2013300}.}
  \label{fig:300w_asm_study_chart}
\end{figure}

\textbf{Ablation Study.}
In Table~\ref{tbl:ASM-improvement-study} we study the ASM assisted loss by calculating the difference between normalized mean errors with and without ASM assisted loss both on ASMNet and MobileNetV2~\cite{sandler2018mobilenetv2}. As shown, using ASMNet utilized with ASM-LOSS function results in 0.96\%, and 1.19\% reduction in NME on 300W~\cite{sagonas2013300}, and WFLW~\cite{wu2018look} respectively. Theses numbers are 0.11\%, and 0.16\% for MobileNetV2~\cite{sandler2018mobilenetv2}. According to the Table~\ref{tbl:ASM-improvement-study}, and Fig.~\ref{fig:300w_asm_study_chart}, using the ASM assisted loss function resulted in more accuracy improvement for ASMNet compared to MobileNetV2~\cite{sandler2018mobilenetv2}. Hence, it can be concluded that the ASM-LOSS function is capable of helping the lightweight CNN much more. In other words, when a lightweight network does not perform accurately enough, using the proposed ASM-LOSS function will play a vital role in improving the performance.

\begin{table}[]
\caption{ Model size (the number of model parameters) and computational cost (FLOPs) analysis of different networks.}
\centering
\small
\resizebox{\columnwidth}{!}{
\label{tbl:size_cost_analysis}
\begin{tabular}{l c c c }
\hline
\multicolumn{1}{l}{Method} & \multicolumn{1}{c}{Backbone} & \multicolumn{1}{c}{\#Params (M)} & \multicolumn{1}{c}{FLOPs (B)} \\ \hline 
DVLN~\cite{wu2017leveraging}                        & VGG-16  & 132.0    & 14.4   \\ 
SAN~\cite{dong2018style}                         & ResNet-152    & 57.4  & 10.7    \\ 
LAB~\cite{wu2018look}                         & Hourglass          & 25.1           & 19.1   \\ 
ResNet50 (Wing + PDB)~\cite{feng2018wing}       & ResNet-50      & 25     & 3.8   \\ 
ASMNet                              & MobileNetV2~\cite{sandler2018mobilenetv2}         & 1.4     & 0.5  \\
MobileNetV2~\cite{sandler2018mobilenetv2}                         &  -     & 2.4      & 0.6        \\ \hline
\end{tabular}
    }
\end{table}

\begin{table}[]
\caption{Investigating the effect of using ASM assisted loss function both on MobileNetV2~\cite{sandler2018mobilenetv2} and ASMNet.}
\label{tbl:ASM-improvement-study}
\centering
\small
\resizebox{6cm}{!}{
\begin{tabular}{l c c c c }
\hline
\multicolumn{1}{l}{\multirow{2}{*}{Method}} & \multicolumn{3}{c}{NME reduction (in \%)} \\
\multicolumn{2}{c}{} & \multicolumn{1}{c}{ASMNet} & \multicolumn{1}{c}{mnv2} \\ \hline
\multirow{3}{*}{300W~\cite{sagonas2013300}} 
& Full & 0.96 & 0.11 \\ 
& Common &  1.58 & 0.05 \\ 
& Challenging & 0.60 & 0.17 \\ 

\hline

\multirow{7}{*}{WFLW~\cite{wu2018look}}
 & Full & 1.19 & 0.16 \\ 
 & Large pose & 0.84 & 0.32 \\
 & Expression & 1.06 & 0.15 \\
 & Illumination & 1.09 & 0.08 \\ 
 & Makeup & 1.29 & 0.25 \\ 
 & Occlusion & 0.13 & 0.90 \\ 
 & Blur & 0.98 & 0.13 \\ \hline
\end{tabular}
}
\end{table}

\textbf{Model Size and Computational Cost Analysis.}
We calculate the number of network parameters as well as FLOPs to evaluate the model size and computational complexity. We calculate the FLOPs over the resolution of $224 \times 224 $. As Table~\ref{tbl:size_cost_analysis}, although ASMNet is the smallest, its performance is comparable with MobileNetV2~\cite{sandler2018mobilenetv2}, one of the best in \textit{compact-class} models. Furthermore, since the idea behind ASMNet is to put a trade-off between accuracy and model performance, as we can see in Table~\ref{tbl:size_cost_analysis}, adding ASM assisted loss to a lightweight model such as ASMNet, and MobileNetV2~\cite{sandler2018mobilenetv2}, results in the accuracy improvement.

\section{Conclusion and Future Work}
\label{sec:conclusion}
In this paper, we proposed ASMNet, a lightweight CNN architecture with multi-task learning for facial landmark points detection and pose estimation. We proposed a loss function that is assisted using ASM~\cite{cootes1995active, ordas2003active} that increases the network accuracy. We built our network (called ASMNet) using a small portion of MobileNetV2~\cite{sandler2018mobilenetv2}. The proposed ASMNet architecture is about 2 times smaller than MobileNetV2~\cite{sandler2018mobilenetv2}, while the accuracy remains at the same rate. The results of evaluating ASMNet and our proposed ASM assisted loss on widely used 300W~\cite{sagonas2013300}, and WFLW~\cite{wu2018look} datasets show that the accuracy of ASMNet is acceptable in detecting facial landmark points and estimating head pose. The proposed method has the potential to be used in other computer vision tasks such as human body joint tracking or other shape objects that can be modeled using ASM. Hence, as a future research direction, we will investigate using ASMNet for such applications.

{\small
\bibliographystyle{ieee}
\bibliography{egbib}
}

\end{document}